\documentclass[10pt,conference]{IEEEtran}
\IEEEoverridecommandlockouts                              

\usepackage{amsmath,amsfonts}
\usepackage{algorithmic}
\usepackage{algorithm}
\usepackage{array}
\usepackage[caption=false,font=normalsize,labelfont=sf,textfont=sf]{subfig}
\usepackage{textcomp}
\usepackage{stfloats}
\usepackage{url}
\usepackage{hyperref} 
\usepackage{verbatim}
\usepackage{graphicx}
\usepackage{booktabs} 
\usepackage{placeins}
\usepackage{cite}
\usepackage{bm}
\usepackage{listings}

\title{\LARGE \bf
Large Language Models for Human-like Autonomous Driving: A Survey}

\author{Yun Li, Kai Katsumata, Ehsan Javanmardi, Manabu Tsukada
\thanks{Graduate School of Information and Science Technology, The University of Tokyo, Tokyo, Japan. Email: \{li-yun, katsumatakai, ejavanmardi, mtsukada\}@g.ecc.e.u-tokyo.ac.jp}
}

\begin{document}

\maketitle
\thispagestyle{empty}

\begin{abstract}

Large Language Models (LLMs), AI models trained on massive text corpora with remarkable language understanding and generation capabilities, are transforming the field of Autonomous Driving (AD). As AD systems evolve from rule-based and optimization-based methods to learning-based techniques like deep reinforcement learning, they are now poised to embrace a third and more advanced category: knowledge-based AD empowered by LLMs. This shift promises to bring AD closer to human-like AD. However, integrating LLMs into AD systems poses challenges in real-time inference, safety assurance, and deployment costs. This survey provides a comprehensive and critical review of recent progress in leveraging LLMs for AD, focusing on their applications in modular AD pipelines and end-to-end AD systems. We highlight key advancements, identify pressing challenges, and propose promising research directions to bridge the gap between LLMs and AD, thereby facilitating the development of more human-like AD systems. The survey first introduces LLMs' key features and common training schemes, then delves into their applications in modular AD pipelines and end-to-end AD, respectively, followed by discussions on open challenges and future directions. Through this in-depth analysis, we aim to provide insights and inspiration for researchers and practitioners working at the intersection of AI and autonomous vehicles, ultimately contributing to safer, smarter, and more human-centric AD technologies.

\end{abstract}


\section{INTRODUCTION}
\begin{figure}
    \centering
    \includegraphics[width=0.5\textwidth]{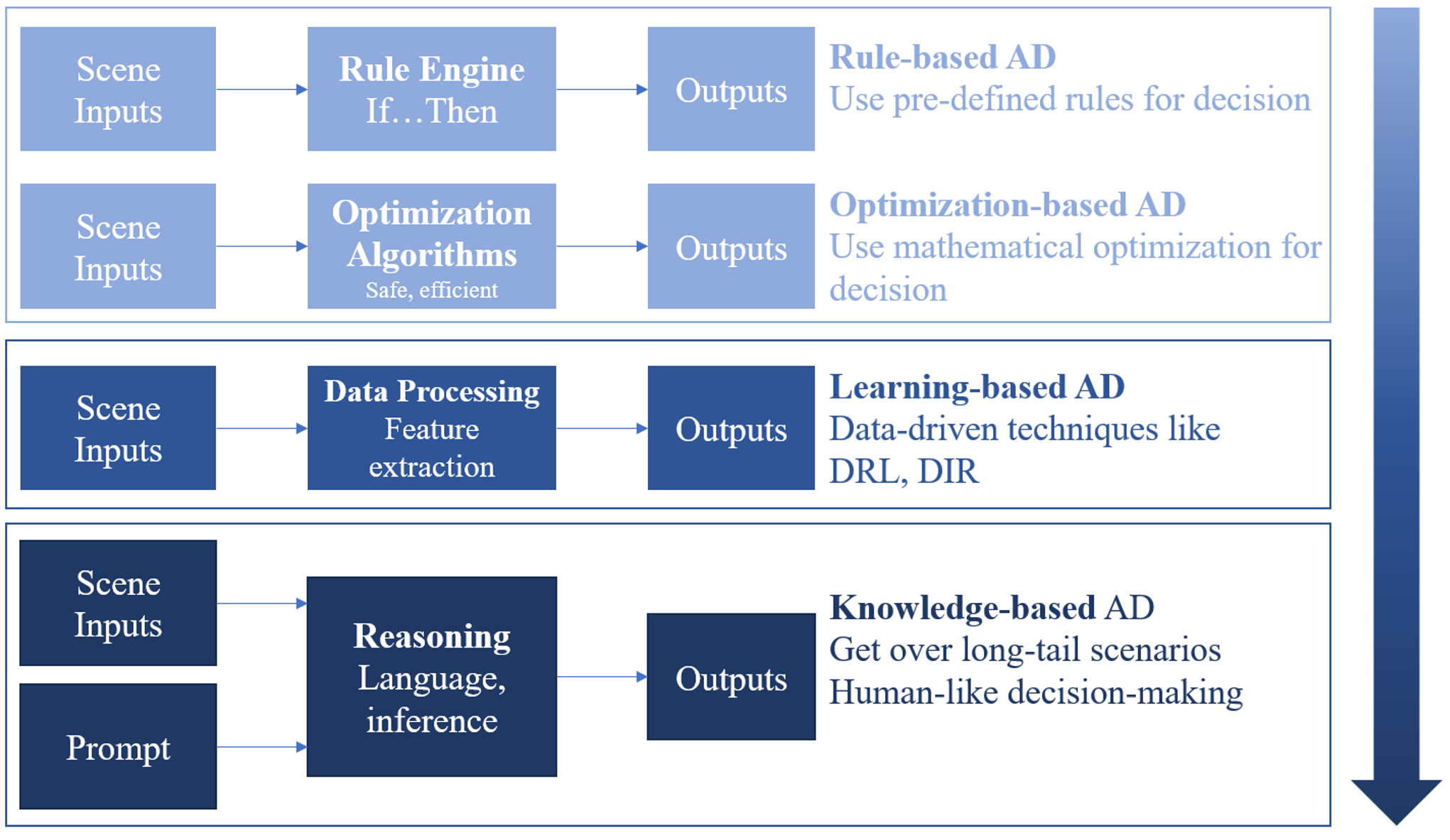}
    \caption{Development of AD systems}
    \label{fig:AD-development}
\end{figure}

Autonomous Driving (AD) has emerged as a transformative technology with the potential to revolutionize intelligent transportation systems, improve road safety, and enhance mobility. At the core of AD lies the decision-making process, which involves analyzing data, understanding the environment, and making informed decisions about navigation and safety. As depicted in Fig. \ref{fig:AD-development}, the development of AD systems can be categorized into three categories based on the techniques employed.

The first category of AD primarily relies on deterministic methods, such as rule-based and optimization-based approaches \cite{Yuan_undated-mi}. While these methods provide reliability and interpretability, they often lack the ability to generalize to complex and novel scenarios \cite{Aksjonov2021-gs, Guanetti2018-mf, Zhao_undated-vo}. The second category of AD, driven by the rapid advancements in deep learning \cite{Silver2016-xs}, leverages techniques such as deep reinforcement learning (DRL) \cite{Aradi2022-fr} in AD systems. Although these learning-based approaches have demonstrated success in handling complex scenarios, they still face challenges in the long-tail scenarios, and as they learn the patterns in the limited training set, they lack generalization in environment out of the training set \cite{Zhao_undated-vo, Wen2024-kg, Kalra2016-se}.

\begin{figure}[t]
    \centering
    \includegraphics[width=0.5\textwidth]{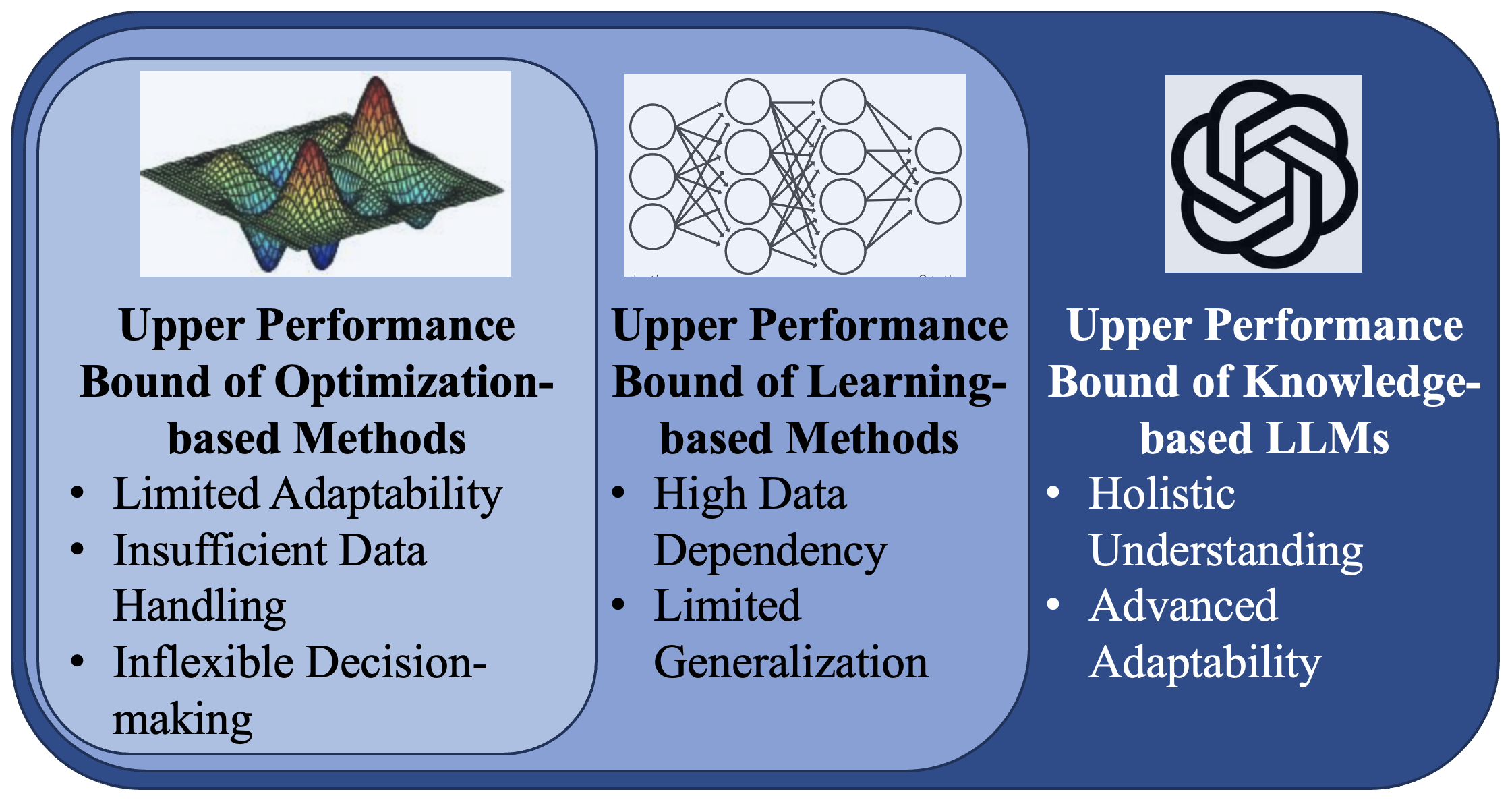}
    \caption{Limitations of conventional methods in AD systems}
    \label{fig:human-like}
\end{figure}

One of the key challenges in AD is addressing the long-tail problem, which refers to the ability to handle rare but complex scenarios in real-world traffic \cite{Feng2023-ot}. Despite efforts to model AD environments using simulators \cite{Zhao_undated-vo, Chen2019-jm, Lu2023-id}, the complexity of real-world scenarios, encompassing factors such as traffic regulations, weather conditions, pedestrian behaviors, and diverse emergencies, poses significant difficulties for existing approaches. Furthermore, the lack of interpretability in AD systems hinders human trust and poses obstacles to their widespread adoption \cite{Zeng2019-yw, Kim2017-se}. Fig. \ref{fig:human-like} intuitively shows the limitations of traditional AD methods, further highlighting the necessity of introducing LLMs.

Recently, the advancement of Large Language Models (LLMs) \cite{Bubeck2023-kx, Wei2022-sx} has ushered in a revolutionary change in AD. Making human-like decisions is crucial for autonomous vehicles as they need to operate among humans in a safe and socially-compliant manner. With their robust language understanding and reasoning capabilities, these models have introduced a new era or third category for autonomous vehicles. On one hand, LLMs can serve as knowledge bases to provide rich commonsense knowledge for AD, compensating for the shortcomings of pure data-driven methods. On the other hand, LLMs are adept at processing natural language instructions, enabling vehicles to understand human intentions and exhibit more human-like driving behaviors. Moreover, LLMs have demonstrated excellent few-shot learning and transfer learning abilities \cite{Wei2022-ta,Brown2020-mz}, which are expected to solve the problems of few corner cases and many long-tail issues in AD. LLMs, represented by GPT-4, have made initial breakthroughs in key tasks such as road risk perception, decision planning, and human-machine interaction, greatly improving the performance of AD. Although there are still many challenges in introducing powerful language models into the complex physical world, LLMs are expected to become a key enabling technology for future AD. AD research sometimes treats LLMs as ``black boxes", neglecting their scientific foundations and general applicability, while AI-oriented research may rely on AD datasets without considering the importance of simulators or real-world testing. This paper aims to bridge these gaps by providing a comprehensive overview of the application of LLMs.

The main contributions of this paper include: 1) systematically reviewing the latest progress of LLMs in the field of AD, and proposing a taxonomy analysis framework covering perception, decision-making, planning, and control; 2) focusing on the two paradigms of modularized decision-making and end-to-end learning, we analyze the advantages and disadvantages of different technical routes in depth, and refine the key scientific problems that need to be tackled next; 3) looking forward to the future development direction and practical deployment needs of LLM-driven autonomous driving, providing references for academia and industry. 

The rest of this paper is structured as follows: Section \ref{par:LLM-intro} delves into the capabilities and foundations of LLMs and MLLMs. Section \ref{par:modulated ad} examines the use of LLMs in modular decision-making for AD, including Table \ref{tab:llm_ad}, which summarizes key studies. Section \ref{par:end-to-end ad} discusses end-to-end LLM-based AD approaches, accompanied by Table \ref{tab:llm_ad_end-to-end} that highlights representative works. Section \ref{par:conclusions} concludes the survey and outlines potential directions for future research.

\section{Large Language Models and Multimodal Large Language Models}
\label{par:LLM-intro}
LLMs and MLLMs have revolutionized the field of natural language processing and beyond, enabling machines to understand, generate, and reason with natural language at an unprecedented level. LLMs are trained on vast amounts of text data, allowing them to capture intricate linguistic patterns, contextual dependencies, and semantic relationships. The input to an LLM is typically represented as a set of tokens $\mathcal{X} = \{\mathbf{x}_{i,j} \mid i \in {1, \dots, N}, j \in {1, \dots, T_i}\}$, where $i$ denotes the $i$-th sentence, $j$ denotes the $j$-th token within the sentence. The output of the LLM, denoted as $\mathbf{y}_{i,j}$, is usually a probability distribution over the next token. Thus, an LLM can be represented as a function $f$ that maps the input to the output:
\begin{equation}
\mathbf{y}_{i,j} = f(\mathbf{x}_{i,j}).
\end{equation}

Building upon the foundation of LLMs, MLLMs have been proposed to handle multimodal inputs, such as images, videos, text, audio, point clouds, and even depth or thermal information \cite{Girdhar2023-pd}. Mathematically, an MLLM can be formulated as a function $f'$ that maps a set of multimodal inputs ${\mathbf{x}_{i,j}^{(1)}, \mathbf{x}_{i,j}^{(2)}, ..., \mathbf{x}_{i,j}^{(n)}}$ to an output $\mathbf{y}_{i,j}$:
\begin{equation}
\mathbf{y}_{i,j} = f'(\mathbf{x}_{i,j}^{(1)}, \mathbf{x}_{i,j}^{(2)}, ..., \mathbf{x}_{i,j}^{(n)}),
\end{equation}
where $\mathbf{x}_{i,j}^{(k)} \in \mathcal{X}^{(k)}$ represents the input from modality $k$ (e.g., image $\mathcal{I}$, text $\mathcal{T}$, point cloud $\mathcal{P}$) for the $(i,j)$-th example, and $\mathbf{y}_{i,j}$ is the model's output (e.g., a control command for AVs) for the same example.

The key components of an MLLM include modality encoders $E^{(k)}$, a multimodal fusion module $F$, a language model $f$, and an output projector $P$. Formally, the MLLM's output can be expressed as:
\begin{equation}
\mathbf{y}_{i,j} = P\left\{f\left[F\left(E^{(1)}(\mathbf{x}_{i,j}^{(1)}), E^{(2)}(\mathbf{x}_{i,j}^{(2)}), ..., E^{(n)}(\mathbf{x}_{i,j}^{(n)})\right)\right]\right\}.
\end{equation}

Several influential MLLMs, such as BLIP-2 \cite{Li2023-pr}, LLaVA \cite{Liu2023-fs}, and Flamingo \cite{Alayrac_undated-lx}, have laid the foundation for the development of this field. These models have demonstrated impressive performance in tasks like Visual Question Answering (VQA), image captioning, and multimodal reasoning. By integrating visual, textual, and potentially other sensory inputs, MLLMs can provide richer contextual information for decision-making processes in complex driving scenarios.
\subsection{Pre-training and Fine-tuning of LLMs and MLLMs}
\subsubsection{Pre-training}
Pre-training plays a pivotal role in the development of LLMs and MLLMs by enabling them to acquire a broad and nuanced understanding of natural language and multimodal data. The mathematical objective of pre-training is described as follows \cite{Radford2018-rc}:
\begin{equation}
\label{equ:pre-training}
\Phi^* = \arg\max_{\Phi} \sum_{i=1}^{N^u} \sum_{j=1}^{T_i} \log P(y_{i,j}^u \mid y_{i,j-c}^u, \ldots, y_{i,j-1}^u, \mathbf{x}_i^u; \Phi),
\end{equation}
where $\Phi^*$ denotes the optimal parameters after pre-training, $N^u$ represents the number of training sequences, and $T_i$ is the length of the $i$-th target sequence. The indices $i$ and $j$ represent the position of the current token in the training data, with $i$ iterating over the training sequences and $j$ iterating over the tokens within each sequence. The context window size $c$ determines the number of previous tokens considered as context when predicting the next token, influencing the model's ability to capture long-range dependencies.
\subsubsection{Fine-tuning}
Fine-tuning is the process of further training a pre-trained LLM or MLLM on specific tasks or domains to enhance its performance. The main fine-tuning techniques include Parameter-Efficient Fine-Tuning (PEFT), prompt tuning, instruction tuning, and Reinforcement Learning from Human Feedback (RLHF).

PEFT methods, such as Low-Rank Adaptation (LoRA) \cite{Hu2021-me}, aim to adapt pre-trained models to specific tasks while minimizing the number of additional parameters and computational resources required. LoRA freezes the pre-trained model weights and injects trainable rank decomposition matrices $\mathbf{A}_i \in \mathbb{R}^{d \times r}$ and $\mathbf{B}_i \in \mathbb{R}^{r \times d}$ into each layer $k$ of the Transformer architecture, where $r \ll d$. The adapted weights $\mathbf{W}'_i \in \mathbb{R}^{d \times d}$ are computed as:
\begin{equation}
\mathbf{W}'_k = \mathbf{W}_k + \mathbf{A}_k\mathbf{B}_k ,
\label{eq:lora}
\end{equation}
where $\mathbf{W}_k \in \mathbb{R}^{d \times d}$ are the frozen pre-trained weights.

Prompt tuning employs learnable continuous vectors called ``soft prompts" \cite{Qin2021-ik} to increase the performance of LLMs in downstream tasks. The objective of prompt tuning is to minimize the following loss function:
\begin{equation}
\mathbf{v}_{1:t}^* = \arg\min_{\mathbf{v}_{1:t}} \left[ -\sum_{i=1}^{N^v} \sum_{j=1}^{T_i} \log P(y_{i,j}^v \mid \mathbf{v}_{1:t}, \mathbf{x}_{i,j}^v) + \lambda \|\mathbf{v}_{1:t}\|_1 \right],
\end{equation}
where $P(y_{i,j}^v \mid \mathbf{v}_{1:t}, \mathbf{x}_{i,j}^v)$ is the probability of generating output $y_{i,j}^v$ given input $\mathbf{x}_{i,j}^v$ and soft prompt $\mathbf{v}_{1:t}$, $N^v$ refers to the number of fine-tuning sequences, and $\lambda \|\mathbf{v}_{1:t}\|_1$ is an $L_1$ regularization term to encourage sparsity in the learned prompts.

Instruction tuning aligns LLMs with human intents by fine-tuning them on a collection of datasets phrased as instructions \cite{Wei2022-ta}. The objective function of instruction tuning can be expressed as:
\begin{equation}
\Phi^* = \arg\max_{\Phi} \sum_{i=1}^{N^v} \sum_{j=1}^{T_i} \log P(y_{i,j}^v \mid x_{i,j}^v, z_{i,j}^v; \Phi),
\end{equation}
where $P(y_{i,j}^v \mid x_{i,j}^v, z_{i,j}^v; \Phi)$ represents the probability of generating the correct output $y_{i,j}^v$ given the input $x_{i,j}^v$ and instruction $z_{i,j}^v$, parameterized by $\Phi$.
RLHF uses human feedback as a reward signal to fine-tune LLMs via RL, generating safer and more aligned outputs \cite{Ouyang2022-fk}. The Boltzmann distribution plays a fundamental role in modeling human choices in the RLHF framework \cite{Jeon2020-yg}:
\begin{equation}
P(l \mid \mathcal{M}(q_{i,j}^v)) = \frac{\exp(\beta \cdot U(l))}{\sum_{l' \in \mathcal{M}(q_{i,j}^v)} \exp(\beta \cdot U(l'))},
\end{equation}
where $\beta \in [0, \infty)$ is a rationality coefficient reflecting the labeler's precision, and $\mathcal{M}(q_{i,j}^v)$ denotes the set of available options for query $q_{i,j}^v$.

Fine-tuning techniques like PEFT and RLHF are particularly beneficial for AD applications. PEFT methods allow for efficient adaptation of large models to specific driving tasks, such as pedestrian behavior prediction or traffic sign recognition, without the need for complete model retraining. RLHF, on the other hand, can be used to align AD systems with human driving preferences and safety standards, potentially leading to more natural and acceptable driving behaviors.
\subsection{In-context Learning and Theoretical Analysis}
In-context learning is a surprising phenomenon that has emerged in LLMs, where the model can be specialized to perform a specific task simply by providing a prompt consisting of a few input-output examples, without any gradient updates to the model parameters \cite{Wies2023-st}. Wies et al. \cite{Wies2023-st} proposed a Probably Approximately Correct (PAC)-learning framework to formalize in-context learning, showing that with high probability, the in-context learner can PAC-learn the downstream task $\tilde{D}$, i.e.,
\begin{equation}
L_{\text{in-context}, \tilde{D}} - \text{Bayes Error Rate} \leq \epsilon,
\end{equation}
where the Bayes Error Rate represents the theoretically lowest possible error rate for any classifier on a given task.
Xie et al. \cite{Xie2022-pi} proposed another theoretical framework where the in-context learning capabilities of LLMs are conceptualized through implicit Bayesian inference. They established several foundational results delineating in-context learning behavior, including asymptotic optimality, monotonic error reduction with example length, and the ability to handle varying-length test examples.

In-context learning enables AD systems to adapt to new scenarios without extensive retraining, crucial for handling diverse and dynamic driving environments. For instance, an LLM-enhanced AD system could quickly adjust its behavior based on a few examples of new traffic patterns or road conditions.

\FloatBarrier

\section{Modular Decision Making}
\label{par:modulated ad}

\begin{table*}[t]
\centering
\caption{Applications of LMs in Autonomous Driving Decision Making}
\label{tab:llm_ad}
\begin{tabular}{@{}p{1cm}p{1.8cm}p{2.5cm}p{3.5cm}p{2cm}p{5.5cm}@{}}
\toprule
Year & LMs Type & Author & Input & Output & Key Contributions \\
\midrule
\multicolumn{6}{l}{\textbf{Autonomous Driving Oriented Approach}} \\
2024/02 & GPT-4 & Azarafza et al. \cite{Azarafza2024-os} & $\mathbf{x}_{\text{img}}, \mathbf{x}_{\text{sensor}}, \mathbf{x}_{\text{reg}}, \mathbf{x}_{\text{weather}}$ & $\mathbf{a}_{\text{brake}}, \mathbf{a}_{\text{throttle}}$ & Enhances LLM reasoning for precise control \\
2023 & GPT-2 & Liu et al. \cite{Liu2023-ij} & $\mathbf{x}_{{\mathbf{s}, \mathbf{a}, r}}$ & $\mathbf{a}_{{\mathbf{s}, \mathbf{a}}}$ & PPO+GPT-2 for safe multi-task decisions \\
2023/12 & gpt-3.5-turbo & Wang et al. \cite{Wang_undated-rw} & $\mathbf{x}_{\text{intent}}, \mathbf{x}_{\text{memory}}$ & $\mathbf{a}_{\text{controller}}$ & LLM selects controllers based on intention \\
2023/09 & GPT-4 & Cui et al. \cite{Cui2023-dt} & $\mathbf{x}_{\text{env}}, \mathbf{x}_{\text{prompt}}$ & $\mathbf{a}$ & Digital twin + LLM in simulation \\
2023/10 & GPT-3.5 & Sha et al. \cite{Sha2023-uq} & $\mathbf{x}_{\text{scene}}, \mathbf{x}_{\text{vehicle}}$ & $\mathbf{a}_{\text{intent}}, \boldsymbol{\theta}$ & LLM+MPC achieves human-like performance \\
2023/09 & gpt-3.5-turbo & Zhang et al.$^\dagger$ \cite{Zhang2024-jr} & $\mathbf{x}_{\text{traffic}}$ & $\mathbf{a}_{\text{suggest}}, \mathbf{a}_{\text{tool}}$ & LLM + traffic analysis tools \\
2023/10 & gpt-3.5-turbo & Cui et al.$^\dagger$ \cite{Cui_undated-fi} & $\mathbf{x}_{\text{traffic}}, \mathbf{x}_{\text{query}}$ & $\mathbf{a}_{\text{high-level}}$ & Real-world LLM testing \\
\midrule
\multicolumn{6}{l}{\textbf{CV/NLP Oriented Approach}} \\
2023 & LLaMA-2, 3.5 & Mao et al.$^\ddagger$ \cite{Mao2023-bt} & $\mathbf{x}_{\text{sensory}}$ & $\mathbf{a}$ & Tool library enhances LLMs \\
2023/12 & GPT-3.5 & Mao et al.$^\dagger$ \cite{Mao2023-qm} & $\mathbf{x}_{\text{scenario}}$ & $\mathbf{a}_{\text{traj}}, \mathbf{a}_{\text{obj}}, \mathbf{a}_{\text{effect}}$ & Prompting-reasoning-finetuning for planning \\
2023/12 & GPT-4 API & Wang et al. \cite{Wang2023-qz} & $\mathbf{x}_{\text{scenario}}$ & $\mathbf{a}_{\text{lane}}$ & LLM-conditioned MPC with feedback \\
2023/09 & gpt-3.5-turbo & Rajvanshi et al. \cite{Rajvanshi2023-xw} & $\mathbf{x}_{\text{graph}}$ & $\mathbf{a}_{\text{nav}}$ & 3D scene graph grounds LLMs \\
2023/12 & LLaMA-2, 3.5, 4\!\!\! & Tanahashi et al. \cite{Tanahashi2023-fj} & $\mathbf{x}_{\text{ego}}, \mathbf{x}_{\text{surrounding}}, \mathbf{x}_{\text{rules}}, \mathbf{x}_{\text{instructions}}$ & $\mathbf{a}_{\text{decision}}, \mathbf{a}_{\text{explanation}}$ & Evaluates LLMs for spatial reasoning \\
\bottomrule
\multicolumn{6}{l}{\footnotesize $^\dagger$Code available. $^\ddagger$Code will be available.}
\end{tabular}
\end{table*}

In this section, we delve into the formulation of AD decision-making using LLMs and explore various approaches to integrate LLMs with traditional AD techniques. We aim to provide a comprehensive overview of state-of-the-art methods and highlight the potential benefits of leveraging LLMs in enhancing vehicle performance, safety, and human-like decision-making capabilities.
AD decision-making can be formulated as a function $F$ that maps input data $\mathbf{x}$ to output actions $\mathbf{a}$, considering the influence of LLMs denoted as $f$:
\begin{equation}
\mathbf{a} = F(\mathbf{x}, f),
\end{equation}
where $\mathbf{x} = \{\mathbf{x}_1, \mathbf{x}_2, ..., \mathbf{x}_n\}$ represents the input data from various sources, such as images, sensor data, traffic regulations, and human queries, and $f$ represents the LLM, which processes the input data and generates output actions $\mathbf{a} = \{\mathbf{a}_1, \mathbf{a}_2, ..., \mathbf{a}_m\}$, including vehicle control, decision-making, and reasoning explanations.
The LLM $f$ can be further decomposed into a pre-trained model $f_{\text{pre}}$ and a fine-tuned model $f_{\text{fine}}$, adapted to the specific AD task:
\begin{equation}
f = f_{\text{fine}}(f_{\text{pre}}, \mathcal{D}),
\end{equation}
where $\mathcal{D}$ represents the domain-specific data used for fine-tuning the pre-trained LLM.
Moreover, the decision-making process can be enhanced by incorporating prompt tuning techniques, denoted as $P$, which guide the LLM to generate more relevant and accurate outputs:
\begin{equation}
\mathbf{a} = F(\mathbf{x}, f, P).
\end{equation}
Considering a traffic scenario with $N$ intelligent agents (vehicles) over a time horizon $T$, the decision-making process can be formulated as:
\begin{equation}
\mathbf{a}_t^i = F(\mathbf{x}_t^i, f, P), \forall i \in {1, 2, ..., N}, \forall t \in {1, 2, ..., T},
\end{equation}
where $\mathbf{a}_t^i$ represents the action of agent $i$ at time $t$, and $\mathbf{x}_t^i$ represents the corresponding input data.

One promising approach for integrating LLMs with AD is to employ these models for precise vehicle control and decision-making in dynamic environments.  Azarafza et al. \cite{Azarafza2024-os} investigated the potential of LLMs in analyzing a combination of images $\mathbf{x}_{\text{img}}$, sensor data $\mathbf{x}_{\text{sensor}}$, traffic regulations $\mathbf{x}_{\text{reg}}$, and weather conditions $\mathbf{x}_{\text{weather}}$ to provide accurate brake $\mathbf{a}_{\text{brake}}$ and throttle $\mathbf{a}_{\text{throttle}}$ control instructions. Similarly, Sha et al. \cite{Sha2023-uq} introduced LanguageMPC, which integrates LLMs with Model Predictive Control (MPC) to handle complex driving scenarios. By providing traffic scene information $\mathbf{x}_{\text{scene}}$ and vehicle states $\mathbf{x}_{\text{vehicle}}$ to the LLM, LanguageMPC generates driving intentions $\mathbf{a}_{\text{intent}}$ and MPC parameters $\boldsymbol{\theta}$ to achieve human-like performance. Wang et al. \cite{Wang_undated-rw} designed a framework that embeds LLMs as a vehicle ``Co-Pilot" to interpret human intention $\mathbf{x}_{\text{intent}}$ and memory $\mathbf{x}_{\text{memory}}$, selecting appropriate controllers $\mathbf{a}_{\text{controller}}$ for specific driving tasks. These findings suggest that LLMs can offer valuable insights for decision-making in complex scenarios, such as low-visibility or long-tail traffic scenarios.

Another approach focuses on integrating LLMs with RL techniques for safe and accurate multi-task decision-making. Liu et al. \cite{Liu2023-ij} proposed the Multi-Task Decision-Making Generative Pre-trained Transformer (MTD-GPT) model, which combines the strengths of RL and GPT to handle multiple driving tasks simultaneously. By training on state-action-reward tuples $\{\mathbf{s}, \mathbf{a}, r\}$ and generating decision-making data sequences $\{\mathbf{s}, \mathbf{a}\}$, MTD-GPT demonstrates superior performance compared to single-task decision-making models.
LLMs have also been combined with digital twin technology to enhance decision-making in simulated environments. Cui et al. \cite{Cui2023-dt} proposed a framework that leverages LLMs' natural language capabilities and contextual understanding to generate actions $\mathbf{a}$ for autonomous vehicles based on environment information $\mathbf{x}_{\text{env}}$ and personalized prompts $\mathbf{x}_{\text{prompt}}$. This framework aims to provide personalized assistance and transparent decision-making.

In the realm of computer vision and natural language processing, Mao et al. \cite{Mao2023-bt} proposed Agent-Driver, which introduces a versatile tool library and cognitive memory to integrate human-like intelligence into autonomous driving systems. By processing sensory data $\mathbf{x}_{\text{sensory}}$, Agent-Driver generates actions $\mathbf{a}$ for autonomous vehicles, demonstrating superior interpretability and few-shot learning ability. Mao et al. \cite{Mao2023-qm} also presented GPT-Driver, which reformulates motion planning as a language modeling problem. By representing scenario descriptions $\mathbf{x}_{\text{scenario}}$ as language tokens, GPT-Driver generates trajectories $\mathbf{a}_{\text{traj}}$, identifies notable objects $\mathbf{a}_{\text{obj}}$, and predicts potential effects $\mathbf{a}_{\text{effect}}$ through a prompting-reasoning-finetuning strategy.

Other notable contributions include TrafficGPT by Zhang et al. \cite{Zhang2024-jr}, which empowers LLMs with the ability to analyze traffic information $\mathbf{x}_{\text{traffic}}$ and provide driving suggestions $\mathbf{a}_{\text{suggest}}$ and tool choices $\mathbf{a}_{\text{tool}}$; DriveLLM by Cui et al. \cite{Cui_undated-fi}, which integrates LLMs with existing autonomous driving stacks for commonsense reasoning in decision-making; SayNav by Rajvanshi et al. \cite{Rajvanshi2023-xw}, which utilizes a 3D scene graph prompt $\mathbf{x}_{\text{graph}}$ to ground LLMs for generating high-level navigation plans $\mathbf{a}_{\text{nav}}$; and the work by Tanahashi et al. \cite{Tanahashi2023-fj}, which evaluates the spatial reasoning and rule-following capabilities of LLMs in autonomous driving scenarios.

These studies highlight the diverse approaches and potential benefits of integrating LLMs into autonomous driving decision-making. From precise vehicle control and multi-task decision-making to personalized assistance and efficient navigation planning, LLMs have demonstrated their ability to enhance the capabilities of autonomous vehicles. As research in this field continues to evolve, the fusion of LLMs with traditional autonomous driving techniques is expected to lead to safer, more efficient, and human-like decision-making processes.

\section{End-to-End Driving}
\label{par:end-to-end ad}

\begin{table*}[t]
\centering
\caption{Applications of LMs in Autonomous Driving End-to-End AD}
\label{tab:llm_ad_end-to-end}
\begin{tabular}{@{}p{0.7cm}p{2.3cm}p{1.8cm}p{2.5cm}p{2.5cm}p{6cm}@{}}
\toprule
Year & LMs Type & Author & Input & Output & Key Contributions \\
\midrule
\multicolumn{6}{l}{\textbf{Autonomous Driving Oriented Approach}} \\
2024/02 & GPT-4, GPT-4V, 3.5 & Fu et al.$^\dagger$ \cite{Fu2024-sm} & $\mathbf{x}_{D_s}, \mathbf{x}_{T_d}, \mathbf{x}_{N_i}$, $\mathbf{x}_{V_c}, \mathbf{x}_{R_n}, \mathbf{x}_{V_i}$ & $\mathbf{a}_{D_d}, \mathbf{a}_T, \mathbf{a}_{C_s}$ & Evaluates (M)LLM-driven AD with memory and reflection \\
2024/01 & N/A & Pan et al.$^\ddagger$ \cite{Pan2024-hk} & $\mathbf{x}_L, \mathbf{x}_B, \mathbf{x}_{T_{ego}}$, $\mathbf{x}_{T_{agents}}$ & $\mathbf{a}_{F_{BEV}}$ & Integrates LLMs with vision-based AD for planning \\
2023/12 & LLaMA-7b & Wang et al. \cite{Wang2023-ag} & $\mathbf{x}_I, \mathbf{x}_P, \mathbf{x}_{T_r}, \mathbf{x}_{S_m}, \mathbf{x}_{U_i}$ & $\mathbf{a}_{D_s}, \mathbf{a}_E$ & Aligns LLM decisions with behavioral planning module \\
2023/10 & GPT-3.5, GPT-4 & Wen et al.$^\dagger$ \cite{Wen2024-kg} & $\mathbf{x}_{V_{sd}}$ & $\mathbf{a}_{A_{acc}}, \mathbf{a}_{A_{id}}, \mathbf{a}_{A_{dec}}$ & Instills knowledge-driven capability into AD \\
2023 & GPT-3.5 & Fu et al. \cite{Fu2023-gb} & $\mathbf{x}_{O_e}$ & $\mathbf{a}_{A_m}$ & Achieves human-like AD with end-to-end LLM \\
2023/09 & GPT-4 & Jin et al. \cite{Jin2023-zq} & $\mathbf{x}_{S_{ego}}, \mathbf{x}_{S_{sur}}$ & $\mathbf{a}_{C_{CARLA}}$ & Combines LLM and CARLA for AD simulation \\
2023/08 & N/A & Tian et al. \cite{Tian2023-yq} & $\mathbf{x}_P, \mathbf{x}_{M_{AD}}$ & $\mathbf{a}_{M_{atom}}$ & Builds modular end-to-end AD with LLMs \\
\midrule
\multicolumn{6}{l}{\textbf{CV/NLP Oriented Approach}} \\
2024/02 & Qwen-VL & Tian et al. \cite{Tian2024-xa} & $\mathbf{x}_{I_s}$ & $\mathbf{a}_{A_m}, \mathbf{a}_D, \mathbf{a}_W$ & Leverages VLM and CoT for hierarchical planning \\
2024/01 & LLaMA2 & Han et al. \cite{Han2024-gg} & $\mathbf{x}_{I_c}, \mathbf{x}_{I_p}, \mathbf{x}_P, \mathbf{x}_{S_v}$ & $\mathbf{a}_{L_h}, \mathbf{a}_{F_a}, \mathbf{a}_{D_s}, \mathbf{a}_R$, $\mathbf{a}_D, \mathbf{a}_{C_s}$ & Emulates interpretable decision making and control \\
2023/10 & gpt-3.5-turbo & Mao et al. \cite{Mao2023-bt} & $\mathbf{x}_S$ & $\mathbf{a}_{T_w}$ & Enhances LLM reasoning with tools and memory \\
2023/02 & Video Swin & Jin et al. \cite{Jin2023-nx} & $\mathbf{x}_{V_n}$ & $\mathbf{a}_{N_a}, \mathbf{a}_{E_r}$ & Trains model for action narration and reasoning \\
2023/10 & LLaMA2 & Xu et al. \cite{Xu2023-tv} & $\mathbf{x}_V, \mathbf{x}_Q, \mathbf{x}_{C_s}$ & $\mathbf{a}_{A_q}, \mathbf{a}_{C_s}$ & Introduces explainable AD with LLM and BDD-X \\
2023/10 & GPT-3.5 & Chen et al. \cite{Chen2023-bm} & $\mathbf{x}_R, \mathbf{x}_{I_v}, \mathbf{x}_{I_p}$ & $\mathbf{a}_A, \mathbf{a}_E$ & Integrates compact numeric vectors into LLMs \\
2023/12 & LLaMA & Shao et al. \cite{Shao2024-tx} & $\mathbf{x}_{D_{CL}}, \mathbf{x}_{I_d}$ & $\mathbf{a}_{C_s}$ & Leverages LLM for closed-loop end-to-end AD \\
2023/12 & BLIP-2 & Sima et al. \cite{Sima2023-ap} & $\mathbf{x}_F$ & $\mathbf{a}_{C_l}$ & Proposes Graph VQA for driving scene reasoning \\
2023/12 & LLaMA-7B & Wang et al. \cite{Wang2023-ag} & $\mathbf{x}_F, \mathbf{x}_{S_m}, \mathbf{x}_{U_i}$ & $\mathbf{a}_{D_l}, \mathbf{a}_E$ & Uses MLLM for closed-loop AD in CARLA \\
\bottomrule
\end{tabular}
\end{table*}

The integration of LLMs into autonomous driving has opened up new possibilities for end-to-end decision-making and control. The end-to-end autonomous driving process with LLMs can be formalized as follows:
$D_t = F(X_t, f, M, P, \Theta), \forall t \in {1, 2, ..., T}$
where $D_t$ represents the driving decision at time step $t$, $X_t$ is the input data (sensor data, navigation information, etc.), $M$ is the memory component, $P$ is the prompt engineering, and $\Theta$ are the learnable parameters. The function $F$ encompasses the entire end-to-end pipeline, from perception to decision-making and control.
To further elaborate on the LLM component $L$, it can be expressed as:
$f = f_{pre} \circ f_{fine} \circ f_{prompt}$
where $f_{pre}$ is the pre-trained LLM, $f_{fine}$ is the fine-tuned model adapted to the autonomous driving domain, and $f_{prompt}$ represents the prompt-tuning techniques employed to guide the LLM's behavior.

Researchers have explored various approaches to integrate LLMs into autonomous driving systems, focusing on different aspects such as closed-loop evaluation, knowledge-driven decision-making, and interpretability.

In the domain of closed-loop evaluation, Fu et al. \cite{Fu2024-sm} introduced LimSim++, a platform for evaluating Multimodal Large Language Model ((M)LLM)-driven autonomous driving. LimSim++ takes scenario descriptions $\mathbf{x}_{D_s}$, task descriptions $\mathbf{x}_{T_d}$, navigation information $\mathbf{x}_{N_i}$, visual content $\mathbf{x}_{V_c}$, road network $\mathbf{x}_{R_n}$, and vehicle information $\mathbf{x}_{V_i}$ as inputs, and generates driving decisions $\mathbf{a}_{D_d}$, trajectories $\mathbf{a}_T$, and control signals $\mathbf{a}_{C_s}$. Similarly, Jin et al. \cite{Jin2023-zq} proposed SurrealDriver, which combines LLMs with the CARLA simulator, taking ego state $\mathbf{x}_{S_{ego}}$ and surrounding state $\mathbf{x}_{S_{sur}}$ as inputs and generating JSON-formatted commands $\mathbf{a}_{C{CARLA}}$ for the CARLA program. Tian et al. \cite{Tian2023-yq} introduced VistaGPT, a framework that uses LLMs to build modular end-to-end autonomous driving systems, taking prompts $\mathbf{x}_P$ and autonomous driving modules $\mathbf{x}_{M_{AD}}$ as inputs and generating atom autonomous driving models $\mathbf{a}_{M_{atom}}$. Wang et al. \cite{Wang2023-ag} proposed DriveMLM, an LLM-based framework that performs closed-loop autonomous driving in realistic simulators, taking figures ($\mathbf{x}_F$), system messages $\mathbf{x}_{S_m}$, and user instructions $\mathbf{x}_{U_i}$ as inputs, and generating low-level decisions $\mathbf{a}_{D_l}$ and explanations $\mathbf{a}_E$. Shao et al. \cite{Shao2024-tx} proposed LMDrive, a language-guided, end-to-end, closed-loop autonomous driving framework that takes camera-LiDAR sensor data $\mathbf{x}_{D_{CL}}$ and driving instructions $\mathbf{x}_{I_d}$ as inputs, and generates vehicle control signals $\mathbf{a}_{C_s}$.

Several studies have focused on integrating LLMs with vision-based autonomous driving frameworks. Pan et al. \cite{Pan2024-hk} proposed VLP (Vision-Language-Planning), which exploits language models to bridge the gap between linguistic understanding and autonomous driving. VLP takes Bird's Eye View (BEV) features, including labels $\mathbf{x}_L$, bounding boxes $\mathbf{x}_B$, ego-vehicle trajectory $\mathbf{x}_{T_{ego}}$, and other agents' trajectories $\mathbf{x}_{T_{agents}}$, and generates enhanced BEV feature representations $\mathbf{a}_{F_{BEV}}$ for improved reasoning and planning. Tian et al. \cite{Tian2024-xa} introduced DriveVLM, which leverages Vision-Language Models (VLMs) and Chain-of-Thought (CoT) for hierarchical planning, taking image sequences $\mathbf{x}_{I_s}$ as input and generating meta-actions $\mathbf{a}_{A_m}$, decisions $\mathbf{a}_D$, and waypoints $\mathbf{a}_W$.

Knowledge-driven decision-making is another area of interest. Wen et al. \cite{Wen2024-kg} proposed DiLu, a framework that instills knowledge-driven capability into autonomous driving using LLMs. DiLu takes scene description vectors $\mathbf{x}_{V_{sd}}$ as input and outputs acceleration $\mathbf{a}_{A_{acc}}$, idle $\mathbf{a}_{A_{id}}$, and decelerate $\mathbf{a}_{A_{dec}}$ actions. Fu et al. \cite{Fu2023-gb} explored the use of LLMs for human-like autonomous driving, taking environment observations 
$\mathbf{x}_{O_e}$and generating discrete meta actions $\mathbf{a}_{A_m}$.

Interpretability and explainability have also been addressed by researchers. Wang et al. \cite{Wang2023-ag} introduced DriveMLM, an LLM-based autonomous driving framework that aligns LLM decisions with the behavioral planning module. DriveMLM takes multi-modal inputs, such as images $\mathbf{x}_I$, point clouds $\mathbf{x}_P$, traffic rules $\mathbf{x}_{T_r}$, system messages $\mathbf{x}_{S_m}$, and user instructions $\mathbf{x}_{U_i}$, and generates decision states $\mathbf{a}_{D_s}$ and explanations $\mathbf{a}_E$. Han et al. \cite{Han2024-gg} proposed DME-Driver, a system that emulates interpretable decision-making and control using a powerful vision-language model, taking current and previous scene images $\mathbf{x}_{I_c}, \mathbf{x}_{I_p}$, prompts $\mathbf{x}_P$, and vehicle status $\mathbf{x}_{S_v}$ as inputs, and generating human-like logic $\mathbf{a}_{L_h}$, focus areas $\mathbf{a}_{F_a}$, scene descriptions $\mathbf{a}_{D_s}$, reasoning $\mathbf{a}_R$, decisions $\mathbf{a}_D$, and control signals $\mathbf{a}_{C_s}$.

Other notable contributions include Agent-Driver by Mao et al. \cite{Mao2023-bt}, which enhances LLM reasoning with tools and memory, taking sensory data $\mathbf{x}_S$ as input and generating trajectories (waypoints) $\mathbf{a}_{T_w}$; ADAPT by Jin et al. \cite{Jin2023-nx}, which trains a model for action narration and reasoning using vehicular navigation videos $\mathbf{x}_{V_n}$, generating action narrations $\mathbf{a}_{N_a}$ and reasoning explanations $\mathbf{a}_{E_r}$; DriveGPT4 by Xu et al. \cite{Xu2023-tv}, an interpretable end-to-end autonomous driving system that takes videos $\mathbf{x}_V$, questions $\mathbf{x}_Q$, and control signals $\mathbf{x}_{C_s}$ as inputs, and generates question answers $\mathbf{a}_{A_q}$ and control signals $\mathbf{a}_{C_s}$; LLM-driver by Chen et al. \cite{Chen2023-bm}, which integrates compact numeric vectors into LLMs, taking route $\mathbf{x}_R$, vehicle $\mathbf{x}_{I_v}$, and pedestrian information $\mathbf{x}_{I_p}$ as inputs, and generating actions $\mathbf{a}_A$ and interpretations $\mathbf{a}_E$; Graph VQA by Sima et al. \cite{Sima2023-ap}, a novel framework for reasoning about a driving scene, taking figures $\mathbf{x}_F$ as input and generating low-level control commands $\mathbf{a}_{C_l}$.

As research in this field progresses, we expect to see more advanced and unified frameworks that leverage the power of LLMs to achieve human-like decision-making and control in autonomous vehicles, addressing challenges such as reasoning, interpretability, and generalization.

\section{Conclusions and Future Directions}
\label{par:conclusions}

This survey comprehensively reviewed the recent progress of integrating LLMs into the knowledge-based AD systems. We first traced the evolution of AD systems, from rule-based and optimization-based approaches to learning-based techniques. We then introduced the key features and training schemes of LLMs that empower them to serve as knowledge bases and reasoning engines for AD. By categorizing existing works into modular AD pipelines and end-to-end AD systems, we analyzed in detail how LLMs can enhance scene understanding, action planning, and human-machine interaction, filling the gap between data-driven AI and human-like AD. Compared with traditional AD methods, LLM-based approaches exhibit superior adaptability to complex scenarios and stronger generalization to novel environments, showing the potential to surpass human drivers in terms of safety and efficiency.

However, the application of LLMs in AD is not without challenges. Firstly, the inference speed and computational cost of LLMs need to be significantly optimized to meet the real-time requirements of AD systems \cite{Abdin2024-xb }. Secondly, the safety, robustness and interpretability of LLM-based decisions must be rigorously validated before deployment in safety-critical AD tasks. Moreover, as LLMs are trained on large-scale online corpora, they may inherit social biases and generate outputs misaligned with human values, raising ethical concerns to be addressed \cite{noauthor_2022-pk}.

To fully harness the potential of LLMs for human-centric AD, we propose the following research directions:
\begin{itemize}
    \item Develop lightweight and efficient LLMs tailored for AD applications with limited computational budgets, e.g., distilling knowledge from large LLMs to smaller models.
    \item Design multi-modal training frameworks that can fuse textual, visual and geographical information to ground LLMs' reasoning in physical AD contexts.  
    \item Introduce safety constraints and ethical guidelines into LLM's training objectives and inference processes to improve the reliability and value alignment of their outputs.
    \item Leverage LLMs to generate human-understandable explanations of AD behaviors to enhance transparency and user trust.
    \item Conduct more closed-loop tests of LLM-based AD in realistic simulators and real-world trials to validate their performance in diverse driving scenarios.
\end{itemize}

In conclusion, the integration of LLMs into AD systems marks an exciting frontier in AI and transportation research. By synergizing the reasoning power of pre-trained models with the learning power of AD agents, this new paradigm holds the promise to achieve human-like or even superhuman vehicle autonomy. We hope this survey can provide a comprehensive reference and inspire more research efforts to address the challenges and realize the full potential of LLMs for human-centric AD.

\bibliographystyle{IEEEtran} 
\bibliography{IEEEabrv,references} 

\end{document}